\documentclass[twocolumn]{article}
\pdfoutput=1
\usepackage[pdftex]{graphicx}

\usepackage{amsmath}
\usepackage{eqparbox}
\usepackage{authblk}
\usepackage{enumitem}
\usepackage{geometry}
\usepackage{todonotes}
\usepackage{cite}
\usepackage{url}
\date{}
\usepackage{color}

\providecommand{\keywords}
{
  \small
  \textbf{\textit{Keywords }}
}
\title{WeiPS: a symmetric fusion model framework for large-scale online learning}

\author{
Xiang Yu, \quad Fuping Chu\thanks{corresponding author}, \quad Junqi Wu, \quad Bo Huang \\
Sina Weibo Inc. \\
\textit{\{yuxiang8, fuping1, junqi5, huangbo2\}@staff.weibo.com}
}

\begin{document}
\maketitle


\begin{abstract}

The recommendation system is an important commercial application of machine learning, where billions of feed views in the information flow every day. In reality, the interaction between user and item usually makes user's interest changing over time, thus many companies (e.g. ByteDance, Baidu, Alibaba, and Weibo) employ online learning as an effective way to quickly capture user interests. However, hundreds of billions of model parameters present online learning with challenges for real-time model deployment. Besides, model stability is another key point for online learning. To this end, we design and implement a symmetric fusion online learning system framework called WeiPS, which integrates model training and model inference. Specifically, WeiPS carries out second level model deployment by streaming update mechanism to satisfy the consistency requirement. Moreover, it uses multi-level fault tolerance and real-time domino degradation to achieve high availability requirement.

\end{abstract}

\keywords Machine Learning, Large-scale Data, Real-time Deploy, Model Stability


\section{Introduction}
\subsection{Online Learning}
The recommendation system is an important commercial application of machine learning, where there are billions of feed views in the information flow every day. Each feed view creates an interaction between user and item, so massive feed views generate large-scale data pose a great challenge to the model training. Thus, so many companies such as ByteDance, Baidu, Alibaba, and Weibo use online learning as an effective way to quickly capture user interests. Also, the frequency of interaction between user and item results in user interests shift quickly. If the interests model cannot be updated in time, the performance of the model will slowly decrease{\cite{he2014practical}}. All in all, real-time online machine learning playing a vital role in the recommendation system.

The overall workflow of online machine learning showed as Figure 1:

\begin{figure}[ht!] 
\centering
\includegraphics[width=3.2in]{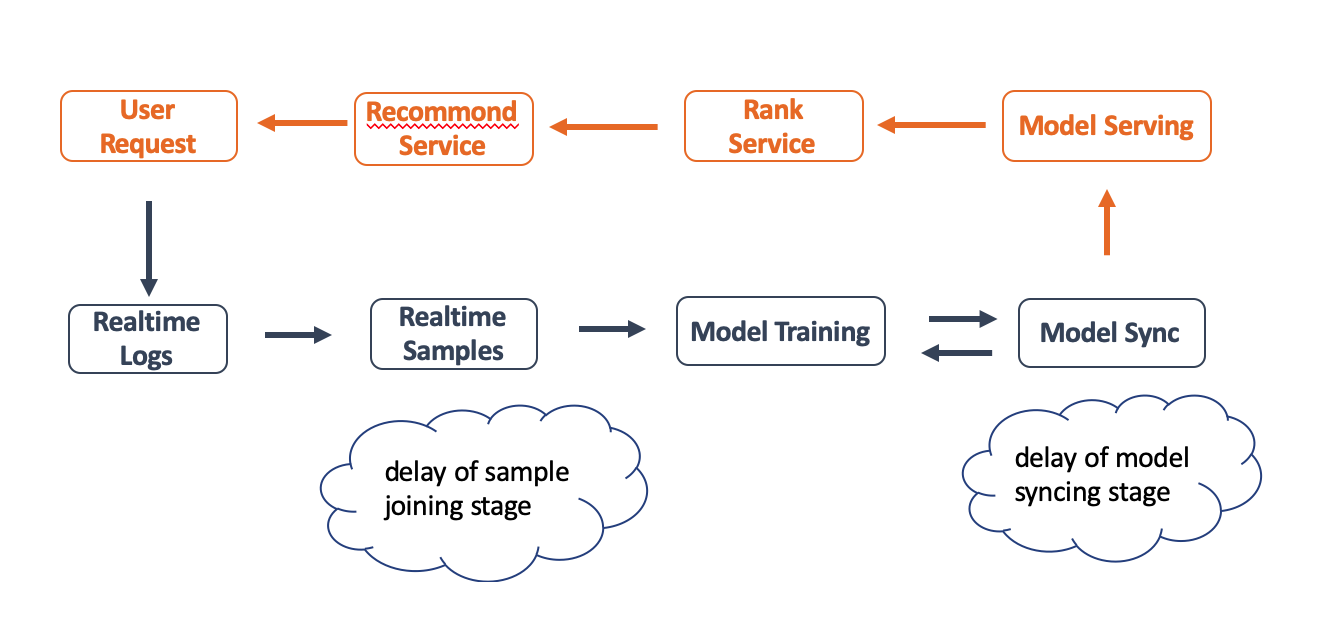}
\caption{ Overall workflow of Online Learning.}
\label{picture}
\end{figure}

The basic process paradigms of online machine learning system are as follows:
\begin{enumerate}[label=\alph*)]
    \item \textbf{Real-time}. This paradigm act on three aspects, sample joining, model training, and parameter synchronization. 1) Real-time samples joining based on user real-time feedback behaviors and real-time exposure data. 2) Real-time model training using the samples to optimize the current model parameters. 3) Real-time parameter synchronization takes a certain update and sharing mechanism to deploy the parameters modified by the model training to the online service environment simultaneously.
    \item \textbf{Online model service}. Combine with real-time parameters synchronization, when a user browsing request comes, the recommendation system uses the latest model parameters for ranking prediction.
\end{enumerate}


\subsection{Read-time Model Deploy}
In the process that user behavior feeding back to the recommendation system of the online learning,  the overall timeliness is mainly limited by two aspects.

Firstly, real-time sample joining exists a certain time window between user exposure and interactive behavior. Online training modules have to wait for this time window during the sample joining so that valid sample data can be spliced. The internet industries have made a trade-off between model effect and timeliness, and also the delay in this part is incomparably avoidable{\cite{li2015click}}. At present, we use Flink{\cite{carbone2015apache}} to support multi-stream sample joining to control service latency at the minutes level.

Secondly, real-time model synchronization, this paper focuses on, refers to the online model service update new model parameters from the model training system to the online model inference system. In most offline scenarios, the model evaluation will be done before the model be deployed to ensure the effectiveness of the new online model, which increases the time interval further. Besides, it is difficult to store hundreds of billions of model parameters in a single machine's memory, so some companies try to use model compression for this case, which results in much greater delay and further affect the timeliness of the model.

We hope parameters change in the model can be updated online in real-time so that user behaviors to be fed back to the online system as soon as possible. However, a series of problems exist in achieving real-time model synchronization between training and prediction. We summarize the actual problems in the production and perform the theoretical analysis, include heterogeneous parameters, heterogeneous requests, sensitivity to system availability, and sensitivity to model stability.

\subsubsection{Heterogeneous Parameters}
In many applications of online and stochastic learning, the input instances are of very high dimension, yet within any model only a few parameters are non-zero. Moreover, according to different purposes, the model parameters can be divided into model parameters for training and model parameters for prediction serving. The parameters required by the two categories will be different in most cases.

In the existing various distributed optimizations, many optimizations need to use auxiliary variables in training, but they are useless for prediction serving. For example, momentum optimization algorithms (Momentum){\cite{sutskever2013importance}}, adaptive learning rate optimization algorithms (Adagrad, RMSProp, Adam){\cite{duchi2011adaptive}}. Furthermore, there are still some inconsistencies between the parameters used in training and prediction in some algorithms, such as Follow The Regularized Leader(FTRL){\cite{mcmahan2011follow}}, which mainly uses N Z during training, and the real parameter W is used in online inference. Additionally, the same inference service corresponding to the same model may have different predictions for various business scenarios. For instance, some need to solve the problem of model service prediction in the ranking service, and some generate features based on the index input by the user. This phenomenon will cause online prediction parameters to be inconsistent with the training parameters. In this case, it is not suitable for a traditional single parameter server cluster.

\subsubsection{Heterogeneous Request}
In the request mode, there are obvious differences between the training stage and the prediction serving stage. The training stage is sensitive to the throughput with a large batch size, and the traffic volume is huge. Relatively, the prediction serving stage is more sensitive to delay time, carry high QPS, set small batch size, and the traffic per request is much smaller.

So there will be differences in the data replica and the model partition mechanism, which requires a fusion framework to customize the cluster division for these two stages.

\subsubsection{Sensitive to System Availability}
In the model training stage, most training engines currently implement cold backup by checkpoint for fault tolerance. Under this mechanism, if the training task is suspended, the model needs to restore from the checkpoint to continue running. For this reason, the service there must be a certain period of unavailability. This fault tolerance mechanism is acceptable in the near-line or the offline system, when the training task hangs on, the newly arrived samples can be cached in the queue waiting for system recovery.

However, the cold backup will cause disasters if it is used in the prediction serving stage. Online prediction serving is called at the moment when a user request occurs, which has a hard limit to the service response time. When the service crash, the recovery delay added by the cold backup fault tolerance makes the online service unavailable, then a large number of user requests fail to return, which is unacceptable for the online system.

\subsubsection{Sensitive to Model Stability}
Online machine learning is different from offline machine learning which has sufficient time to evaluate the model and determine whether the model can be used for online production. Online machine learning, especially in the case of updates the model in seconds, is difficult to guarantee the stability and the effectiveness of the model while ensuring the timeliness of the model. Besides, the offline evaluation mechanism is no longer suitable for online machine learning, so new evaluation methods and model rollback mechanisms are needed to ensure the stability of online business effects.

We have discussed the basic process properties of the online machine learning system in detail. To satisfy above properties, we introduce a symmetric fusion framework based on large-scale online machine learning. Next, we will illustrate our design and implementation.
\begin{enumerate}[label=\alph*)]
    \item We propose a symmetric fusion industrial system framework for large-scale online learning called WeiPS, which integrates training and prediction.
    \item We implement a streaming synchronization mechanism that can speed up the model synchronization period to second level. Moreover, this mechanism is under the satisfy of eventual data consistency.
    \item We also complete multi-level fault tolerance and downgrade mechanism to meet the high availability requirements in different situations.
\end{enumerate}

\begin{figure*}[!htp] 
\centering
\includegraphics[width=6.2in]{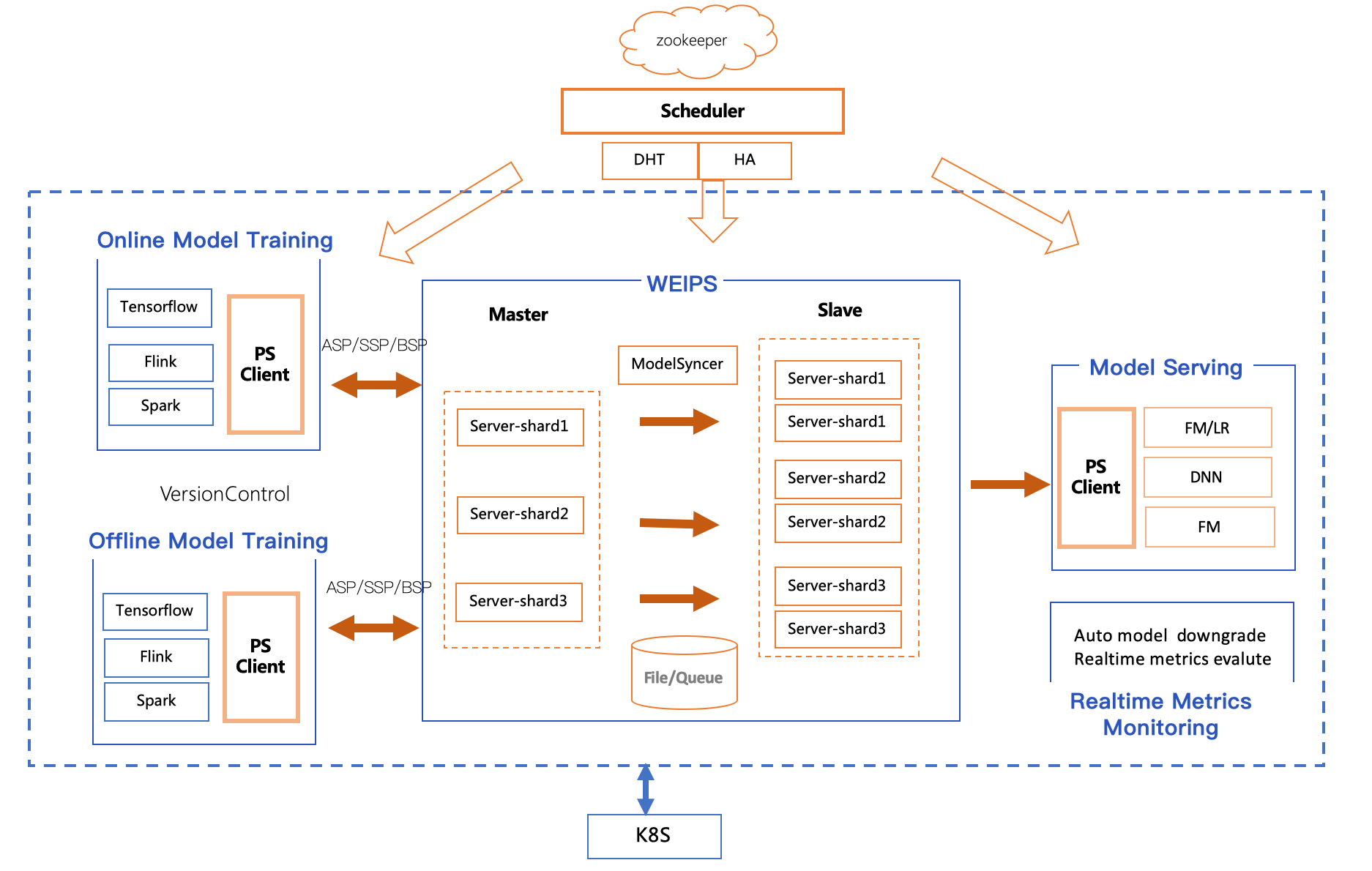}
\caption{ Architecture of WeiPS}
\label{picture}
\end{figure*}

\section{Related work}
\subsection{Large-scale Machine Learning}
Parameter server(PS) framework, proposed by Mu Li, is designed for solving distributed machine learning problems{\cite{li2014scaling}}. The server nodes maintain globally shared parameters that were represented as dense or sparse vectors and matrices. The framework manages asynchronous data communication between nodes and supports flexible consistency models, elastic scalability, and continuous fault tolerance. Many parts of our framework refer to this architecture.

HBM-PS proposes a hierarchical workflow that utilizes GPU High-Bandwidth Memory, CPU main memory, and SSD as 3-layer hierarchical storage{\cite{zhao2020distributed}}. All the neural network training computations are contained in GPUs. HBM-PS Make full use of the advantages of modern hardware to solve the problems encountered in large-scale sparse deep learning.

\subsection{Online Learning}

ODL paper presents a new model structure framework and a new optimization algorithm that attempts to tackle the challenges by learning DNN models which dynamically adapt depth from a sequence of training data in the online learning setting{\cite{sahoo2017online}}.

XDL	framework provides a series of mechanisms for online learning, such as feature entry filter, incremental model export, features expire, and so on. XDL has a practical effect on online learning, and WeiPS get many useful proposes from XDL{\cite{jiang2019xdl}}.

\subsection{Data synchronization}
As the parameter server is similar to the key-value distributed storage, we survey common open source systems, such as redis{\cite{redis}}, Tair{\cite{tair}}, Cassandra{\cite{cassandra}} and so on. Although all of them have many advanced replication and data synchronization mechanisms, they do not meet the needs of model synchronization in machine learning scenarios, such as master nodes and slave nodes have different data structures and computing modes.


\section{Architecture}
At present, most of the works in the industry focus on the real-time training of large-scale models, and there is relatively little research on the effectiveness of large-scale model real-time inference and model deployment. Meanwhile, these two parts are precisely the important roles of online machine learning. WeiPS is designed for the fusion of training and inference, which seamlessly integrating multiple training frameworks and inference frameworks. It also has automatic monitoring indicators and a downgraded system. The entire architecture of WeiPS is depicted in Figure 2.

WeiPS follows the principle of module cohesion and symmetric, and its architecture is divided into three system roles named worker, server, and scheduler. The worker has two different categories, trainer and predictor, and the server also has two types, master and slave.

\subsection{Worker}
The worker is responsible for feature processing and model training.
\begin{itemize}
    \item The trainer is responsible for large-scale sample training of the model.
    \item The predictor is responsible for online high-performance model prediction service.
\end{itemize}
The interactions between the servers are all through WeiPS-client. It is worth mentioning that because the predictor and the trainer have different scheme requirements, WeiPS-client carries different characteristics for that.

\subsection{Server}
The server is responsible for the update of the gradients and the storage of model parameters. The server cluster provides a unified access interface externally and uses the sharding mechanism internally to perform multi-node fragmentation of parameters. As the training and the prediction scenarios have different requirements for high availability, the slave and the master will adopt different distributed fault-tolerant architectures.

\begin{itemize}
    \item The master mainly interacts with the trainer cluster, and it uses a checkpoint mechanism for cold backup of fault tolerance.
    \item The slave mainly interacts with the predictor cluster, and it uses a multi-replica mechanism for fault tolerance and load balancing.
\end{itemize}

Also, to accomplish the deployment and update of the online learning model in seconds, the server use a streaming synchronization mechanism to update the model parameters, which we will demonstrate in detail subsequently.

\subsection{Scheduler}
The scheduler is the core scheduling component of the entire cluster, which is responsible for the lifecycle management of the entire system and other components, as well as the management of data consistency. The scheduler component maintains global metadata and is stateless. The guarantee of metadata consistency are managed by the open-source consistency coordination system (such as ZooKeeper{\cite{hunt2010zookeeper}}, ETCD{\cite{etcd.io}}). In this way, the robustness and maintainability of the entire cluster are greatly improved.

Besides, all the components of the WeiPS architecture are deployed and orchestrated based on K8S{\cite{burns2016borg}}, using K8S's powerful service management capabilities and docker's{\cite{merkel2014docker}} virtualization capabilities not only improve the stability and scalability of overall system services but also promote the efficiency of maintenance and deployment.


\section{Implementation}

To improve the real-time performance of online learning models, we implement second level deployment and update mechanisms for large-scale models.

\subsection{Streaming synchronization}

\begin{figure*}[!htp] 
\centering
\includegraphics[width=6.2in]{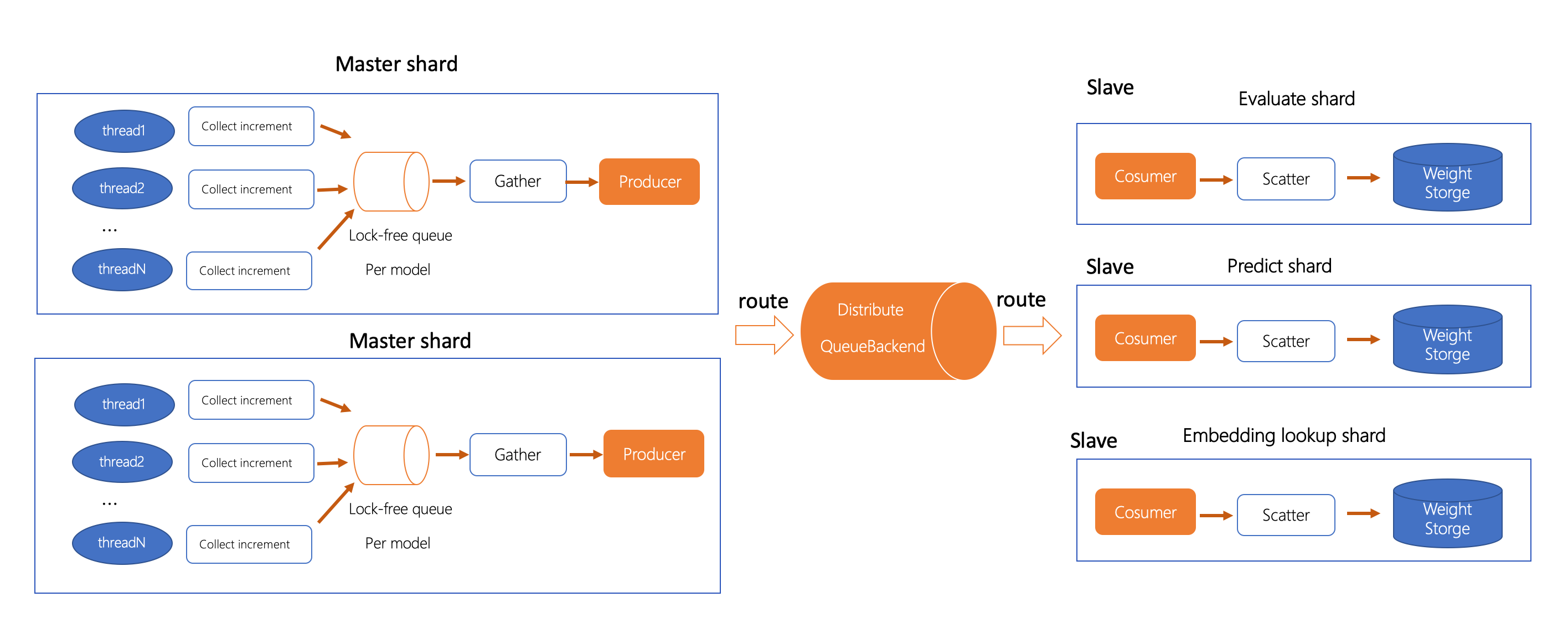}
\caption{ Mechanism of Streaming Synchronization}
\label{picture}
\end{figure*}

WeiPS proposes a streaming synchronization mechanism that can raise the timeliness of model updating and solve the problem of model heterogeneity.
Streaming synchronization uses checkpoint for full synchronization between replicas, and incremental real-time synchronization through external queues. We refer to the master-slave synchronization mechanism of Redis, but there are many differences to support machine learning business scenarios, WeiPS's master-slave synchronization has the following characteristics:

\begin{enumerate}[label=\alph*)]
    \item Asynchronization synchronous: Distributed external queues are introduced between the master and slave to synchronize data asynchronously to increase system throughput. Master and slave perform flow decoupling to prevent data synchronization from affecting model training or model prediction.
    \item Data transform: WeiPS slave is not simply a data copy for the Master, it will perform corresponding data screening and data conversion according to the type of slave, and the final output data can satisfy model evaluation, model prediction, or other embedding queries, etc. Business functions, through module encapsulation of these differences, adapt to the future expansion of new model use scenarios.
    \item Feature Filter: For controlling the effective size of the model, online learning generally offers a feature filtering mechanism to clean up model parameters that are no longer used in time, that can save model space and improve model generalization performance. This requires real-time synchronization to support parameter deletion.
    \item Eventual consistency: To ensure the idempotence and final consistency of queue consumption and production data, simplify the failure mechanism and reduce network traffic and fault tolerance costs, the synchronization parameters between the WeiPS master and slave are the increments of the ID granularity. If an ID changes during the synchronization cycle, the external queue will push the full amount of this ID, not only the current and last time increment of this ID.
\end{enumerate}

There are four types of update mechanism: collect, gather, push, scatter. Figure 3 summarizes how these mechanisms collaborate. Next, we will illustrate how these mechanisms work.
\subsubsection{Collector}
After receiving the push request from the client, the model collects the parameters in real-time and then writes them to the internal lock-free cache queue. To save memory space for the sparse model, the data collected at this time only include the collection ids and the operation type. This procedure does not retain the model increment. Also, we use the lock-free queue to collect the weight increment generated in the multi-threading to ensure thread safety without affecting the parameter update performance.

\subsubsection{Gather}
The gather reads the incremental index from the lock-free queue, and aggregate the corresponding updates,  then generate the final data that needs to be sent according to the logic of gather. Here, we abstract two important issues that are aggregation frequency and aggregation method. To meet various business needs and flexible facilitate experiments and selection of businesses, the gathering frequency part supports three modes:
\begin{enumerate}[label=\alph*)]
    \item Real-time gather: Real-time aggregation is the most time-efficient model synchronization method, but it requires high network bandwidth. Through experiments, we observed that the repetition rate of model parameters updates within 10 seconds reach 90\% or much more, which also provides a basis for subsequent bandwidth optimization based on gathering methods.
    \item Threshold-based gather: The model will be aggregated and updated when the increment parameters reach a certain threshold.
    \item Period-based gather: This mode accumulates parameters at certain intervals, which will be sent to the external queue.
\end{enumerate}

Data gathering is implemented in a model-related manner because different model or optimizer has different storage manner and usage scenario. For example, LR-FTRL has 3 sparse matrices, and FM-FTRL has 6 sparse matrices. FM-SGD has two sparse matrices, and DNN is generally multiple spare matrices plus multiple dense matrices.

\subsubsection{Pusher}
The pusher takes care of pushing parameters from master to producer of Kafka. Before sending the model parameters to the external queue, we make serialize and compress for the aggregated updated data. Especially, we combine the concept of fragmentation of the external queue with the fragmentation mechanism of the Parameter Server.  So the model parameters sent by each master node will be stored in a specific partition of the distribute queue through performing the partition mapping according to the server-id before sending.

\subsubsection{Scatter}
The scatter is responsible for consuming model parameters from the external queue used by the slave. Also, the slave can specify certain partitions for consuming so that there is no need to read the full Kafka queue while reducing bandwidth pressure. Just like the master, the slave also handles a server-id to partition-id routing mapping methods, for example,  modulo operation.
Each shard obtains the corresponding model parameters through the shard routing, and then the scatter performs a summary and updates to the local parameter memory storage.
We make two important improvements, flexible model routing and model transformer to solve heterogeneous traffic and heterogeneous traffic problems that we have mentioned in Section 1.
\begin{enumerate}[label=\alph*)]
     \item Model Routing: The traffic of model training is inconsistent with the traffic of model inference in the production environment. This means the resource requirements of the two situations is inconsistent, which corresponding to WeiPS demand the number of shards will be different. Through the router mechanism, the master and the slave can update the real-time model even the shards are inconsistent. For the same slave role, the number of shards of different slaves is diverse, which greatly improves the flexibility of business application and decline cluster costs.
    \item Model Transforming: Due to different business scenarios, real-time updates will face the problem of heterogeneous master-slave data which requires real-time model conversion during the real-time synchronization process. This part strongly depends on the model type so we offer a high-level abstraction for model-related operations.
\end{enumerate}

\begin{figure}[ht!] 
\centering
\includegraphics[width=3.2in]{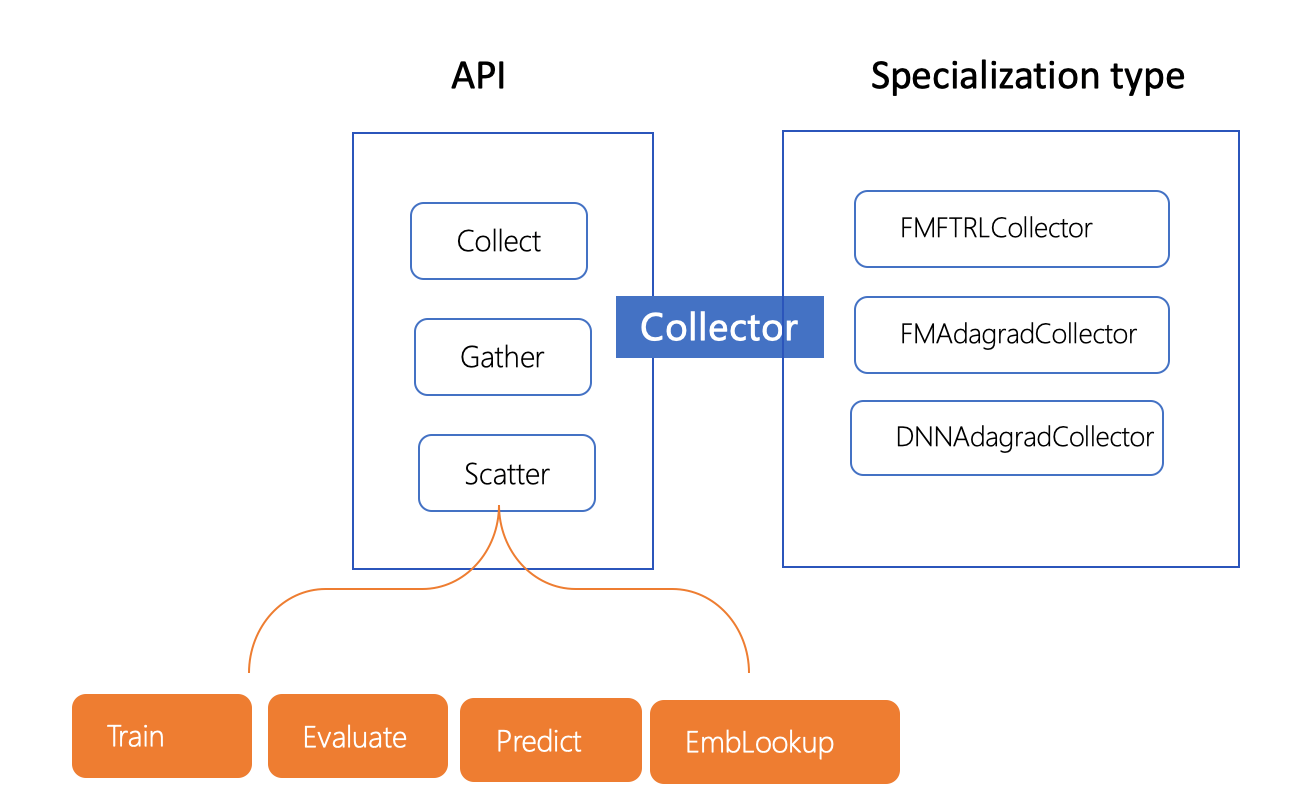}
\caption{ Types of Collector And Scatter }
\label{picture}
\end{figure}

\subsection{High Availability and Reliability}
The consequence of business indicators is directly affected by WeiPS system stability. The WeiPS parameter service cluster is the center of distributed model sharing and optimization in the machine learning application architectures. To meet the stability requirement of different architectures at the same time, WeiPS currently supports a multi-level availability fault tolerance guarantee, which is a cold backup and hot backup by multi-replica.

\subsubsection{Cold Backup of Fault Tolerance}
This is mainly used in the offline model training, which corresponding to the master role in the WeiPS architecture. Cold backup achieves its goal by the checkpoint as same as the saving checkpoint principles of the traditional parameter service training architecture,  and it can be divided into two steps: 1) \textbf{Saving Checkpoint}. The scheduler sends save checkpoint requests to each master shard node periodically, and each node stores the model parameters according to the received request information and saves them in the disk. 2) \textbf{Loading Checkpoint}. When the daemon service quit unexpectedly, the model parameters are read from the disk and loaded into a different master shard.

The cold backup mechanism of WeiPS is different from traditional architecture since WeiPS is an online service architecture and the same WeiPS cluster will serve multiple versions of multiple models at the same time, including model training, model evaluation, model prediction, etc. Therefore, we design five expand features for cold backup fault tolerance mechanism to meet business requirements:
\renewcommand{\labelenumi}{\roman{enumi}}
\begin{enumerate}[label=\alph*)]
    \item We use the random trigger and asynchronous saving mechanisms to prevent traffic aggregation through the model saving checkpoint process.
    \item We design several patterns to solving the different business requirements for data consistency. Firstly, due to the writing speed of remote storage is lower than local disk, we use a hierarchical storage strategy that the local disk backup time interval can be less than an hour and the remote storage time interval is generally at the hour or day level. Secondly, we use an external queue as the real-time incremental backup to achieve strong consistency. 
    \item Customization model backup mechanism and configuration of service hot switch allow different fault tolerance strategies(including local save interval, remote save interval, whether turn on incremental backup, etc.). At the same time, as the running stage of the same model, the fault tolerance strategy can also be dynamically switched according to the actual demands.
    \item The dynamic routing function in the model loading process can make failover migration between heterogeneous clusters easily. For example, if the model owner wants to migrate a model from cluster A has 10 shards to cluster B has 20 shards, WeiPS can automatically mapping all data slices.
    \item WeiPS also support partial fault tolerance to shortening the recovery time. When an individual node crashes down, the entire cluster will not be restarted, and only this shard will recover from the replica checkpoint.
\end{enumerate}

\subsubsection{Hot Backup For Fault Tolerance}
Hot standby fault tolerance corresponds to a multi-replica load balancing mechanism. This mechanism is mainly used in the online service scenario with high availability requirement. When an instance of the online service node crashes,  the other instance takes over the requests that belong to that node.

\begin{figure}[ht!] 
\centering
\includegraphics[width=3.2in]{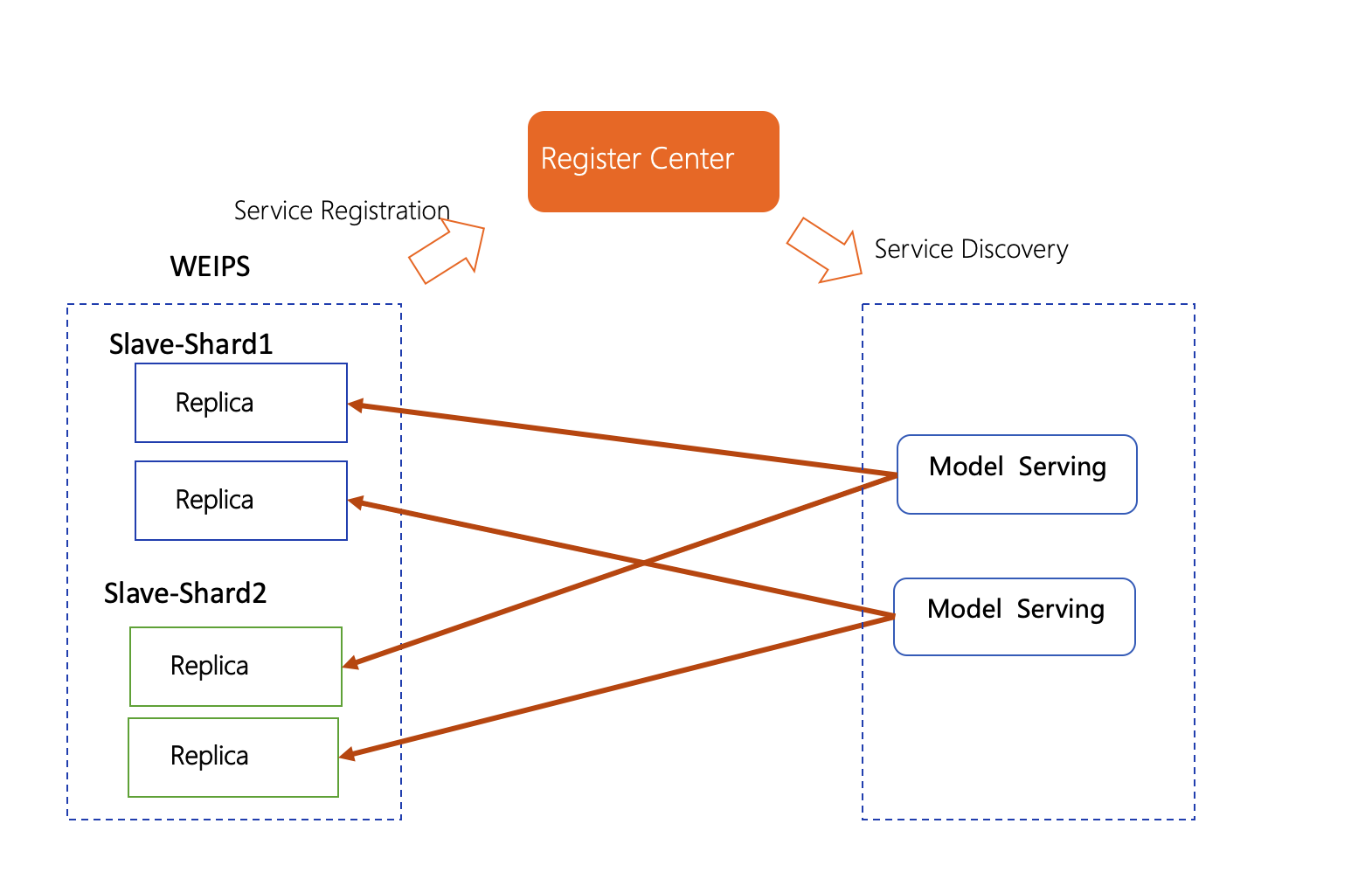}
\caption{ Multi-replica of Load Balancing }
\label{picture}
\end{figure}
The principle of multi-replica of load balancing is similar to the service discovery mechanism and it's a relatively mature solution. However, the general load balancing is mainly for stateless services,  and online learning as a stateful service, so load balancing must solve the problem of data consistency between replicas. The solutions are full synchronization and streaming incremental synchronization that just mentioned above.

\subsection{Monitor And Downgrade}
Although streaming synchronization can maximize the advantages of effectiveness, it also introduces instability to the online model effect. To solve the problems caused by the second level update, the real-time monitoring module and the domino downgrade automatic version switching function are crucial.

\subsubsection{Model Metrics Monitoring}
There are two problems with traditional model evaluation:
\begin{enumerate}[label=\alph*)]
    \item The evaluation data is not real-time data. The data distribution of the content recommendation scenario changes rapidly and the samples used in the evaluation are offline samples. So the evaluation metrics on these samples are not sufficiently convincing.
    \item Evaluation samples can't participate in model training. Even with samples produced in real-time, traditional evaluation methods are unable to update the model with the verification samples. As a result, the user feedback information corresponding to these samples are lost, which will have a certain impact on low frequency users or new items.
\end{enumerate}
Accordingly, WeiPS uses the predicted result of the training samples as the estimated result of the current model parameters, this happens before the training sample data update gradients. According to the prediction, WeiPS calculates the objective function results and actually update gradients. Through this mode, we can guarantee the real-time performance and integrity of the samples.



\subsubsection{Domino downgrade}
The entire domino downgrade process is divided into two segments, downgrade trigger, and downgrade execution.
The downgrade here refers to recover the model to the previous latest stable version when the model occurs an abnormal change. Throughout the external service process of the server node, the server node will periodically generate new checkpoints and use the newest checkpoint as the model version. Furthermore, the offset address of the external queue at that time will be saved in the checkpoint to support the streaming update.

\begin{enumerate}[label=\alph*)]
    \item \textbf{Downgrade trigger.} The simplest way is to set a threshold for the indicator, and it can be triggered when the threshold is reached. But this may occur false alarms in action. At this time, a smoothing threshold strategy that sample a few more contrast points can be used, and the threshold after smoothing can better catch the true change of the data distribution.
    \item \textbf{Downgrade execution.} WeiPS supports the hot switching between model versions. When the downgrade is triggered, WeiPS pick an appropriate version as the target version according to the indicator of the historical version. Then it sets the target version as the latest version to completes the downgrade process.
\end{enumerate}
Generally, the downgrade trigger and downgrade execution are extraordinarily flexible that the person can specify the appropriate version and opportune moment for switching manually. Similarly, it also can automatically downgrade according to the version switching strategy. The version switching strategy includes the latest version strategy and the optimal index version strategy.



\section{Conclusion}
Real-time model deployment and model stability are critical to online learning. To deal with the challenge of model parameters synchronization, we introduce WeiPS, which proposes a second level streaming update mechanism to ensure parameter synchronization between training server and serving server. Multi-level fault tolerance and real-time domino degradation mechanisms guarantee the stability of the model service.

WeiPS is still evolving, which is being improved in many aspects, for example, (1) supporting reinforcement learning, (2) introducing distributed hash table(DHT) to support dynamic cluster scale-out and scale-in, (3) providing more consistent checkpoint for fault tolerance.


\bibliographystyle{unsrt}
\bibliography{WEIPS}

\end{document}